\documentclass{article}
\usepackage{graphicx} 
\usepackage{float}
\usepackage{subfig}
\usepackage{url}
\usepackage{setspace}
\usepackage[margin=1in]{geometry}

\begin{document}

\begin{titlepage}
    \centering
    \vspace*{2.5cm}
    
    {\LARGE \bfseries Design and Implementation of an OCR-Powered Pipeline\\[0.3cm]
    for Table Extraction from Invoices\par}
    
    \vspace{2cm}

    {\Large Parshva D. Patel\\[0.3cm]
    \texttt{parshvapatel211@gmail.com}\par}

    \vfill
    
    {\large July 2025}

\end{titlepage}

\begin{abstract}
Invoices often contain structured information such as sender details and product tables, but real-world invoice processing remains challenging when the data comes from scanned or photographed images rather than digital PDFs. This paper presents a robust system for automated invoice data extraction using a hybrid pipeline that combines OpenCV-based pre-processing with OCR and advanced table extraction techniques. Our approach addresses real-world challenges including skewed perspectives, variable lighting, noise from signatures, barcodes, staplers, and broken table structures. We segment invoices into detail and product sections, apply hybrid table detection using both Img2Table and manual fallback methods, and finally generate structured JSON outputs using row-wise OCR. This method proves particularly effective for physical invoices with multiple products and complex layouts, significantly reducing the need for manual data entry.
\end{abstract}
\vspace{1.5cm} 
\section{Introduction}

In the era of digital transformation, businesses and organizations increasingly rely on automated systems to manage operational tasks efficiently. Among these, invoice processing remains a critical function across industries—from retail and logistics to finance and supply chain management. While many large enterprises employ advanced enterprise resource planning (ERP) systems, a significant portion of small and medium businesses (SMBs) continue to deal with physical or scanned invoices. Manual entry of such data not only introduces errors but also consumes considerable human effort and time.
\vspace{0.5cm}
Although modern Optical Character Recognition (OCR) tools and PDF extractors are highly effective for clean, digitally-generated invoices, their performance declines significantly in real-world scenarios. Invoices photographed via mobile devices are often affected by perspective distortion, uneven lighting, handwritten marks, stamps, barcodes, stapler holes, and other physical artifacts. These inconsistencies pose serious challenges to accurate table detection and data extraction—particularly when dealing with multi-line items and variable layouts.
\vspace{0.5cm}
This report presents the design and implementation of a robust, image-processing and OCR-powered pipeline aimed at solving these challenges. By integrating traditional computer vision techniques from OpenCV with adaptive pre-processing logic and third-party libraries such as Img2Table and Tesseract OCR, we develop a hybrid system capable of handling unstructured and noisy invoice images.
\vspace{0.5cm}
The proposed system performs the following tasks:
\begin{itemize}
    \item Perspective correction and skew adjustment for tilted or misaligned images
    \item Pre-processing to eliminate barcodes, noise, and non-textual artifacts
    \item Table structure recognition in both bordered and borderless layouts
    \item Extraction and mapping of key-value pairs (e.g., HSN, Sr. No., Product, Amount) into structured JSON format
\end{itemize}
The focus of this work is particularly on invoices containing multiple line items, sub-tables, and sender–receiver metadata, where existing solutions tend to fail due to inconsistent layouts and lack of semantic clues. Through this internship project, we aim to demonstrate a generalized, scalable pipeline that improves data extraction accuracy and robustness across diverse invoice formats.

\newpage
\section{Related Work}

Invoice data extraction has been a key area of research in the domain of document understanding, OCR, and intelligent automation. Numerous techniques—ranging from traditional rule-based pipelines to modern deep learning frameworks—have been proposed and adopted in both academic and industrial settings.
\vspace{0.5cm}
Open-source Optical Character Recognition (OCR) engines such as \textbf{Tesseract}, \textbf{EasyOCR}, and \textbf{PaddleOCR} have been widely utilized for text extraction from images and scanned documents. While these tools perform well on high-resolution images or clean scans, their effectiveness decreases considerably when faced with poor lighting, skewed angles, watermarking, or physical noise such as stapler holes and handwriting artifacts. Additionally, OCR engines alone do not inherently understand document structure or relationships between extracted elements such as headers and values.
\vspace{0.5cm}
Commercial tools like \textbf{Adobe Scan}, \textbf{ABBYY FineReader}, and \textbf{Google Document AI} offer robust document processing capabilities, especially for digital PDFs or standardized formats. However, their performance drops on unstructured invoice images commonly encountered in the real world. Moreover, these solutions are often proprietary, expensive, and act as black boxes—making them less suitable for flexible customization in specialized use cases like invoice photos from mobile devices.
\vspace{0.5cm}
For structured table extraction, tools like \textbf{img2table} have gained traction. This open-source library provides table structure detection and content parsing from PDFs and images. While it excels with bordered and neatly aligned tables, its accuracy declines in cases of broken, multi-column layouts, or skewed images. Our work builds upon img2table and enhances it with pre-processing, image corrections, and logic-based fallback techniques for improved reliability.
\vspace{0.5cm}
Deep learning-based methods have also been explored for document layout analysis and field extraction. Architectures such as \textbf{YOLOv5}, \textbf{Detectron2}, and \textbf{TableNet} have been trained for detecting text regions, tables, and semantic blocks in documents. These models show high accuracy on benchmark datasets but demand large-scale annotated data for effective training. Moreover, their generalization across diverse invoice layouts and languages is still limited. Deployment of such models often incurs heavy compute and maintenance costs, which are impractical for lightweight, real-world applications.
\vspace{0.5cm}
Hybrid approaches, which combine traditional computer vision methods with rule-based logic and OCR, have shown promise in handling a wide variety of document types. These methods often leverage techniques like adaptive thresholding, edge detection, contour hierarchy analysis, and heuristic filtering to isolate and extract useful information without deep model dependence.
\vspace{0.5cm}
In summary, existing solutions tend to perform well under constrained or standardized conditions but falter in scenarios involving noisy, unstructured invoice images. Our work contributes a lightweight yet robust hybrid pipeline that addresses this gap—bridging the advantages of classical image processing, OCR technologies, and practical domain heuristics to support real-world invoice parsing from mobile-captured photos.

\newpage
\section{Methodology}

\subsection{Image Preprocessing and Perspective Correction}
Invoices captured using mobile devices or low-quality scanners often exhibit skew, perspective distortion, and edge artifacts. These visual defects significantly degrade the performance of OCR and downstream table structure detection.

To address these issues, we developed a preprocessing pipeline that performs the following:

\begin{itemize}
\item \textbf{Edge Cropping:} A fixed 10-pixel strip is removed from all four sides of the image to eliminate peripheral noise and unnecessary border artifacts. This helps prevent false detections during contour extraction.

\item \textbf{Contour Filtering:} We apply OpenCV's \texttt{findContours} to identify all contours in the binarized image. Contours with area less than 35,000 pixels are discarded to remove small artifacts like signatures, logos, or noise blobs. The largest valid contour is selected as the document boundary.

\item \textbf{Corner Extraction:} Using the filtered contour, we extract the convex hull and identify the four extreme corner points. These are reordered (top-left, top-right, bottom-right, bottom-left) based on geometric heuristics to maintain consistency.

\item \textbf{Perspective Correction:} The identified corners are passed to \texttt {cv2.getPerspectiveTransform()} 
and warped using 
\texttt{cv2.warpPerspective()}. This transforms the image to a flat, rectangular, top-down view, correcting tilt and skew.

\item \textbf{Padding:} A 4-pixel white border is added after warping to prevent edge-cutting in later operations like dilation and OCR cropping.

\end{itemize}

This module ensures the document is well-aligned and preprocessed for optimal downstream analysis, improving both visual clarity and text recognition quality.

\subsection{Document Noise Handling}
Scanned invoices and mobile captures often contain undesired visual elements such as stapler pins, punch holes, folds, smudges, or shadows. These introduce artificial contours and break up line continuity, hampering table extraction and OCR.

Our noise handling pipeline includes:

\begin{itemize}
\item \textbf{Adaptive Thresholding:} We apply \texttt{cv2.adaptiveThreshold()} to binarize the image while adapting to local brightness variations, enhancing text clarity and minimizing the effect of uneven lighting.

\item \textbf{Morphological Dilation:} Dilation reconnects broken edges caused by noise or light shadows. We use a \(3 \times 3\) rectangular kernel iteratively to consolidate text regions and lines, increasing structural coherence.

\item \textbf{Artifact Tolerance:} While some small noise may persist, the robust morphology and contour-filtering ensure only clean rectangular shapes are used for structure detection, allowing us to extract usable table data even from suboptimal scans.

\end{itemize}

This step is particularly critical for ensuring consistency across a diverse range of invoice sources.

\subsection{Signature and Tick Artifact Removal}
Invoices often contain handwritten checkmarks, signatures, or scribbles added during manual verification. These artifacts—especially when in blue or black ink—may be mistakenly detected as text or table boundaries by the OCR or structure-detection module.

Our solution involves the following multi-step approach:

\begin{itemize}
\item \textbf{Color Segmentation:} We convert the image to HSV color space and create a mask targeting blue and cyan hues, typically associated with pen markings.

\item \textbf{Edge Detection:} We apply Canny edge detection to the color-segmented region to isolate irregular curves and loops characteristic of handwritten content.

\item \textbf{Morphological Filtering:} Small-area, irregular, and thin contours that don’t resemble text or table lines are identified and isolated.

\item \textbf{Masking and Removal:} The detected regions are filled with white pixels to eliminate them, ensuring that the structure of the invoice—especially tables and headers—remains clean.

\end{itemize}

This step significantly improves OCR precision by preventing misclassification of artifacts as invoice content.

\subsection{Barcode Detection and Removal}
Many invoices feature barcodes, especially in e-commerce and logistics sectors. These barcodes contain densely packed vertical lines that confuse OCR engines and may be misinterpreted as tables or characters.

Our pipeline includes:

\begin{itemize}
\item \textbf{Gaussian Blurring:} A Gaussian blur (typically \(5 \times 5\)) is applied to smooth barcode patterns, reducing edge sharpness without compromising text clarity.

\item \textbf{Rectangular Dilation:} A vertical rectangular kernel is used to combine tightly packed barcode lines into a solid black region for easier detection.

\item \textbf{Contour Detection:} Contours with a high aspect ratio (height/width ratio) and uniform fill intensity are flagged as barcodes.

\item \textbf{Conditional Removal:} Depending on user preference, these regions are either retained (if barcode data is needed) or masked with white pixels to prevent interference with text extraction.

\end{itemize}

This module enhances the visual clarity of the document and prevents incorrect segmentation in later stages.

\subsection{Section-wise Table Segmentation Using Keywords and Spatial Rules}
Invoices often have two visually and logically distinct table regions:

\begin{itemize}
\item \textbf{Top Section:} Contains metadata such as invoice number, date, sender and buyer details, and tax information.
\item \textbf{Bottom Section:} Contains line items with product descriptions, HSN codes, quantities, pricing, and tax breakdown.
\end{itemize}

To separate these:

\begin{itemize}
\item \textbf{Keyword Detection:} We use OCR and fuzzy string matching to detect anchor words like "Invoice No", "Buyer", "HSN", and "Total".

\item \textbf{Spatial Grouping:} Detected keywords are grouped based on their Y-axis location. We apply a distance threshold to distinguish between closely spaced vs. distant headers.

\item \textbf{Region Cropping:} Horizontal image slices are created using bounding boxes between detected keywords to isolate each section for individual processing.

\item \textbf{Dedicated Pipelines:} Each cropped section is routed to a separate OCR pipeline—one optimized for key-value headers, the other for tabular data.

\end{itemize}

This allows more precise handling of each section's unique layout.

\subsection{Line Removal for OCR Enhancement}
Invoice tables frequently include gridlines which interfere with OCR by segmenting words and numbers across cell boundaries.

Our line-removal strategy involves:

\begin{itemize}
\item \textbf{Preprocessing:} The image is converted to grayscale, binarized, and inverted.

\item \textbf{Vertical and Horizontal Masking:} Using morphological operations with long \(1 \times n\) and \(n \times 1\) kernels, we extract vertical and horizontal lines separately.

\item \textbf{Combined Line Mask:} The individual masks are combined and then subtracted from the inverted binary image to remove the gridlines.

\item \textbf{Cleanup:} An opening operation with a small circular kernel removes the leftover noise and dots.

\end{itemize}

Removing gridlines improves OCR accuracy by reducing character fragmentation.

\subsection{Invoice Header Extraction via Hybrid Approach}
Invoice headers vary in structure, language, and formatting, making it challenging to extract consistent key-value pairs.

We designed a three-layer fallback system:

\subsubsection{Primary Approach – img2table}
We use the open-source \texttt{img2table} library to detect tabular structures in the header. Tables are assigned unique IDs and filtered using size and bounding-box overlap heuristics to eliminate duplicates.

\subsubsection{Fallback – Manual OpenCV Detection}
For images where \texttt{img2table} fails, we apply edge detection and morphology to find boxes and text-like regions. The detected bounding boxes are filtered by size and grouped into rows to mimic key value structures.

\subsubsection{Final Fallback – Row-wise Extraction}
If no tabular structure is detected, the image is processed row-wise. After line removal, OCR is applied across text lines, and field-value pairs are extracted using colon delimiters and regular expressions.

\subsection{Product Details Extraction – Bounding Box + OCR}
This step parses the product table containing itemized rows.

\begin{itemize}
\item \textbf{Input:} A clean image of the lower tabular region is provided from the previous segmentation stage.

\item \textbf{Dilation:} We dilate characters inside cells to merge words and digits into block-like regions.

\item \textbf{Contour Detection:} Bounding boxes are drawn around merged text blobs.

\item \textbf{Filtering:} Contours are filtered based on height and aspect ratio to remove isolated dots and fragments.

\item \textbf{Row Sorting:} Contours are sorted by Y-axis to reconstruct table rows. Within each row, boxes are sorted left-to-right for column alignment.

\item \textbf{OCR Extraction:} Tesseract OCR is run line-by-line. The result is post-processed to correct for splitting or merging errors and matched to header keys for consistent structuring.

\end{itemize}

This hybrid OCR-contour strategy ensures high accuracy even with misaligned or loosely structured invoices.
\newpage
\section{Implementation and Techniques}

This section describes the practical implementation of the invoice processing pipeline, including the software environment, tool configurations, and architectural decisions made during development.

\subsection{Development Environment}
The project was developed in Python 3.10 using Jupyter notebooks and modular scripts. Key libraries used include:
\begin{itemize}
    \item \textbf{OpenCV (v4.9)} for image preprocessing, contour analysis, and inpainting.
    \item \textbf{Tesseract OCR (v5.3)} via \texttt{pytesseract}, with \texttt{--psm 6} and \texttt{--oem 3} settings for structured layout reading.
    \item \textbf{img2table} for layout-based table extraction and conversion to structured DataFrames.
    \item \textbf{NumPy} and \textbf{Pandas} for data handling and transformation.
\end{itemize}

\subsection{Modular Architecture}
The implementation follows a modular pipeline with the following independent stages:
\begin{enumerate}
    \item \textbf{Preprocessing Module:} Handles grayscale conversion, edge detection, contour extraction, and perspective correction.
    \item \textbf{Noise Removal Module:} Removes signatures, barcodes, and ticks using color thresholding and inpainting.
    \item \textbf{OCR and Parsing Module:} Uses Tesseract to extract text from preprocessed image regions (headers, body, tables).
    \item \textbf{Table Detection Module:} Uses \textbf{img2table} to detect and extract structured rows/columns.
    \item \textbf{Key-Value Structuring Module:} Matches header labels with standard invoice fields using exact and fuzzy logic.
\end{enumerate}

\subsection{Integration of \texttt{img2table}}
The \texttt{img2table} library was selected for its robustness in handling:
\begin{itemize}
    \item Variably aligned or irregular tables.
    \item Mixed text and numeric content across columns.
    \item Output as structured Pandas DataFrames without manual threshold tuning.
\end{itemize}
It was applied after line removal and OCR cleanup to ensure only relevant text was processed.

\subsection{Custom Configurations and Observations}
\begin{itemize}
    \item OCR worked best with \texttt{--psm 6} (assumes uniform block layout).
    \item Signature and barcode removal significantly reduced false positives in extracted text.
    \item \texttt{img2table} required fine-tuning image resolution; images were resized to a minimum of 800px width before processing.
\end{itemize}

\newpage

\section{Challenges and Solutions}

During the development of the document processing pipeline, several real-world challenges were encountered, primarily due to the variability in invoice formats, scanning conditions, and physical document artifacts. This section outlines the most significant challenges and the corresponding techniques implemented to address them.

\subsection{Holes, Punches, and Page Damage}

Invoices often contain punch holes or physical defects that appear as sharp white or black regions, interfering with contour detection. These were effectively handled using a combination of Canny edge detection and morphological dilation. By expanding the detected edges, such artifacts were either suppressed or isolated from meaningful document structures.

\subsection{Uneven Lighting and Multiple Light Sources}

Images captured under inconsistent lighting conditions exhibited brightness gradients and shadow regions. To mitigate this, all input images were converted to a standardized format using grayscale normalization and histogram equalization. This enhanced contrast and eliminated lighting-induced inconsistencies, improving the reliability of thresholding and OCR.

\subsection{Irregular or Nested Table Structures}

Some invoices presented irregular table layouts, such as nested or unbordered cells, breaking the standard tabular assumptions. To address this, we integrated the \texttt{img2table} library, which leverages OCR-aware heuristics and spatial clustering to detect structured table blocks even when borders are missing or misaligned.

\subsection{Inconsistent Threshold Parameters}

A single set of binarization or morphological parameters could not generalize across varying invoice designs and qualities. To solve this, we introduced a dynamic preprocessing pipeline where multiple thresholding methods (Otsu, adaptive Gaussian, etc.) were conditionally applied based on image entropy and contrast. This approach ensured robustness across both dark-text and faint-text regions.

\subsection{Overlapping Handwritten Marks and Stamps}

User-added elements such as signatures, handwritten ticks, or ink stamps frequently obscured critical printed text. These were identified using their color dominance or shape irregularity (non-text blobs) and masked via inpainting techniques. This preserved the underlying layout for OCR while discarding non-printed noise.

\subsection{Perspective Distortion and Document Skew}

Captured images were often not properly aligned with the scanner or camera frame, resulting in geometric distortions. A contour-based four-point perspective transformation was applied to rectify the orientation, ensuring that text lines and tables aligned horizontally and vertically, thus improving OCR performance.

\subsection{Header and Key-Value Misalignment}

In multi-column invoices, headers such as \texttt{HSN}, \texttt{Amount}, or \texttt{Description} may not align perfectly with their values. A row-wise segmentation technique was used to ensure each line item’s header and value remained logically associated. Column clustering using vertical projection histograms helped resolve minor alignment errors.

\subsection{Generalization Across Templates}

Invoices from different vendors vary significantly in layout, fonts, and language cues. To generalize across these variations, a modular pipeline was adopted with clear separation between preprocessing, layout detection, OCR, and post-processing. This allowed individual components to be tuned independently and replaced if necessary.

\paragraph{Summary.}  
These challenges reflect real-world document variability and highlight the need for adaptive, modular processing pipelines. By combining classical image processing with robust heuristics and OCR integration, the proposed system demonstrates resilience across a wide range of invoice conditions.

\section{Results}

\begin{figure}[H]
\centering
\includegraphics[width=\textwidth,height=0.8\textheight,keepaspectratio]{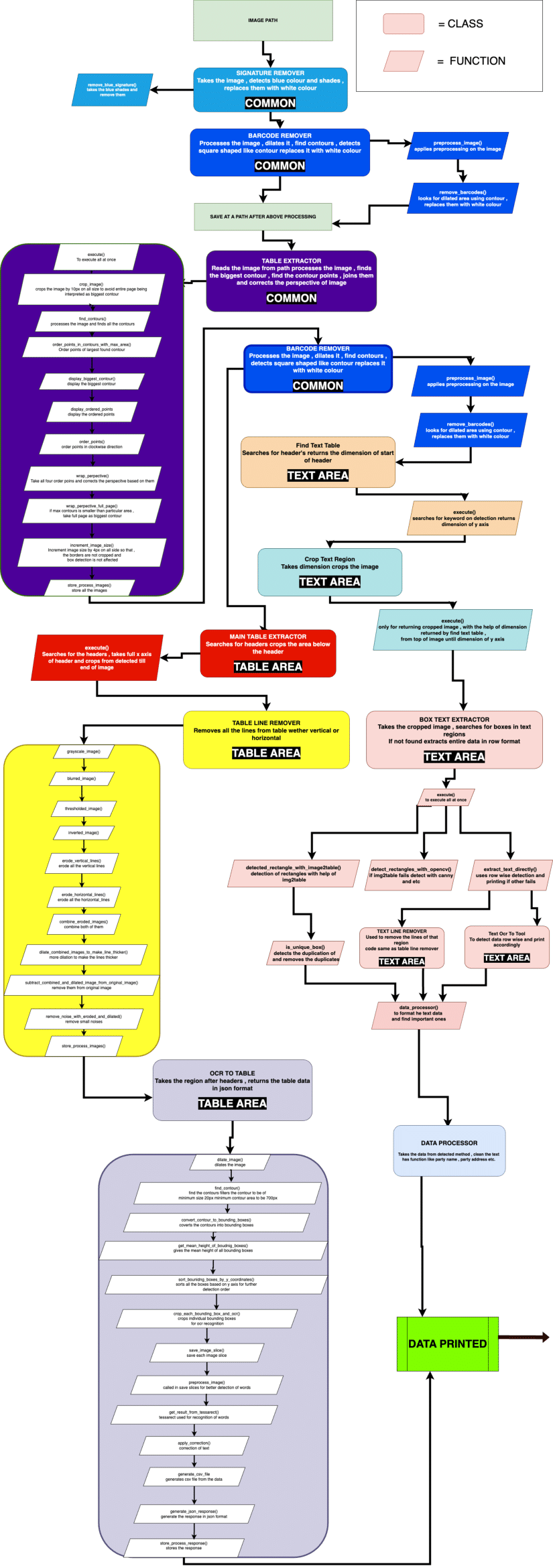}
\caption{Overview of the Document Processing Pipeline}
\label{fig:flowchart}
\end{figure}

This section showcases the visual output and intermediate stages of the document processing pipeline. Each sub-section corresponds to a distinct processing module used in the extraction of structured data from invoice images.

\subsection{Image Preprocessing and Perspective Correction}

\begin{figure}[H]
\centering
\subfloat[Original Image]{\includegraphics[width=0.3\textwidth]{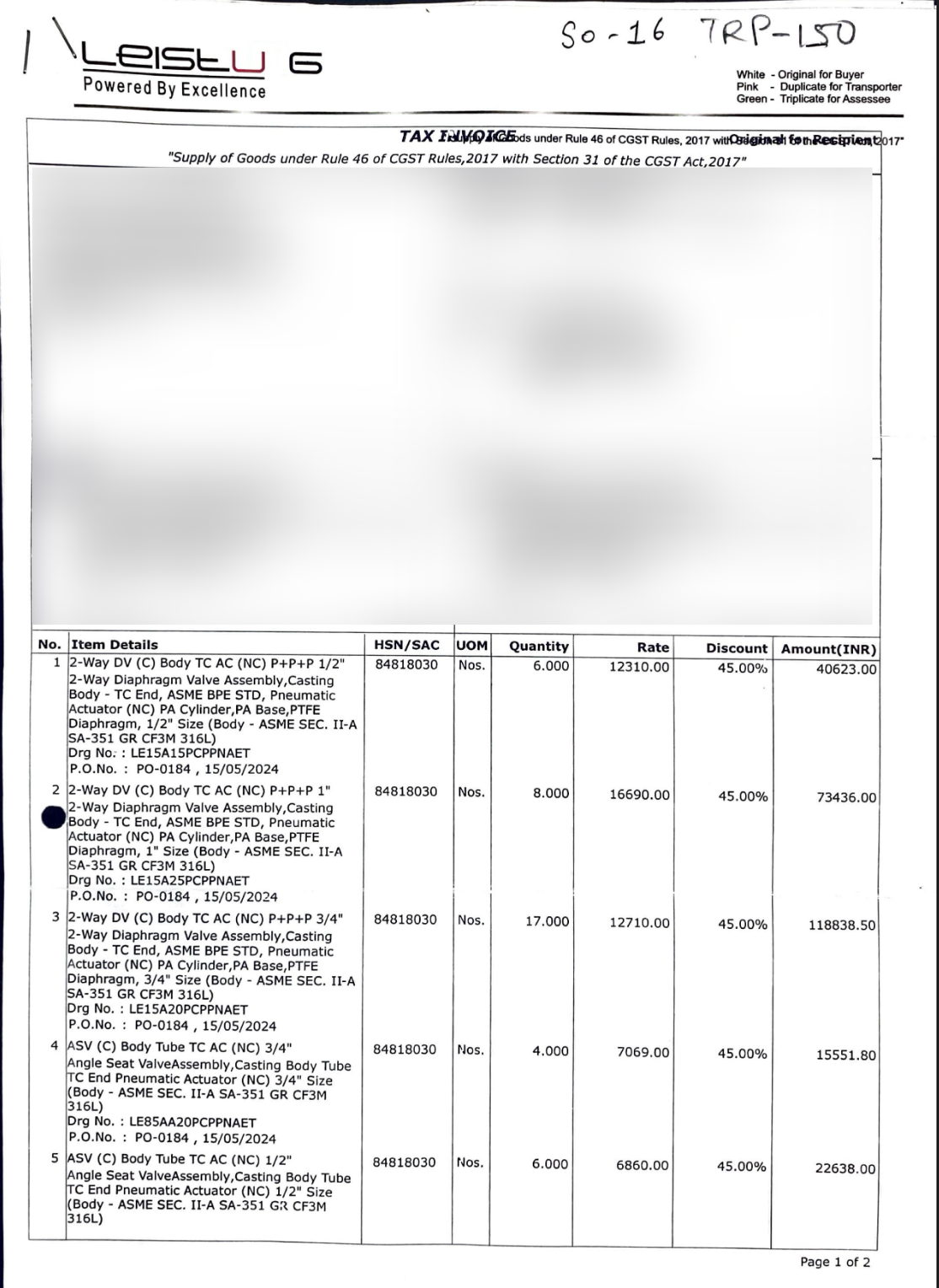}}
\hfill
\subfloat[All Contours]{\includegraphics[width=0.3\textwidth]{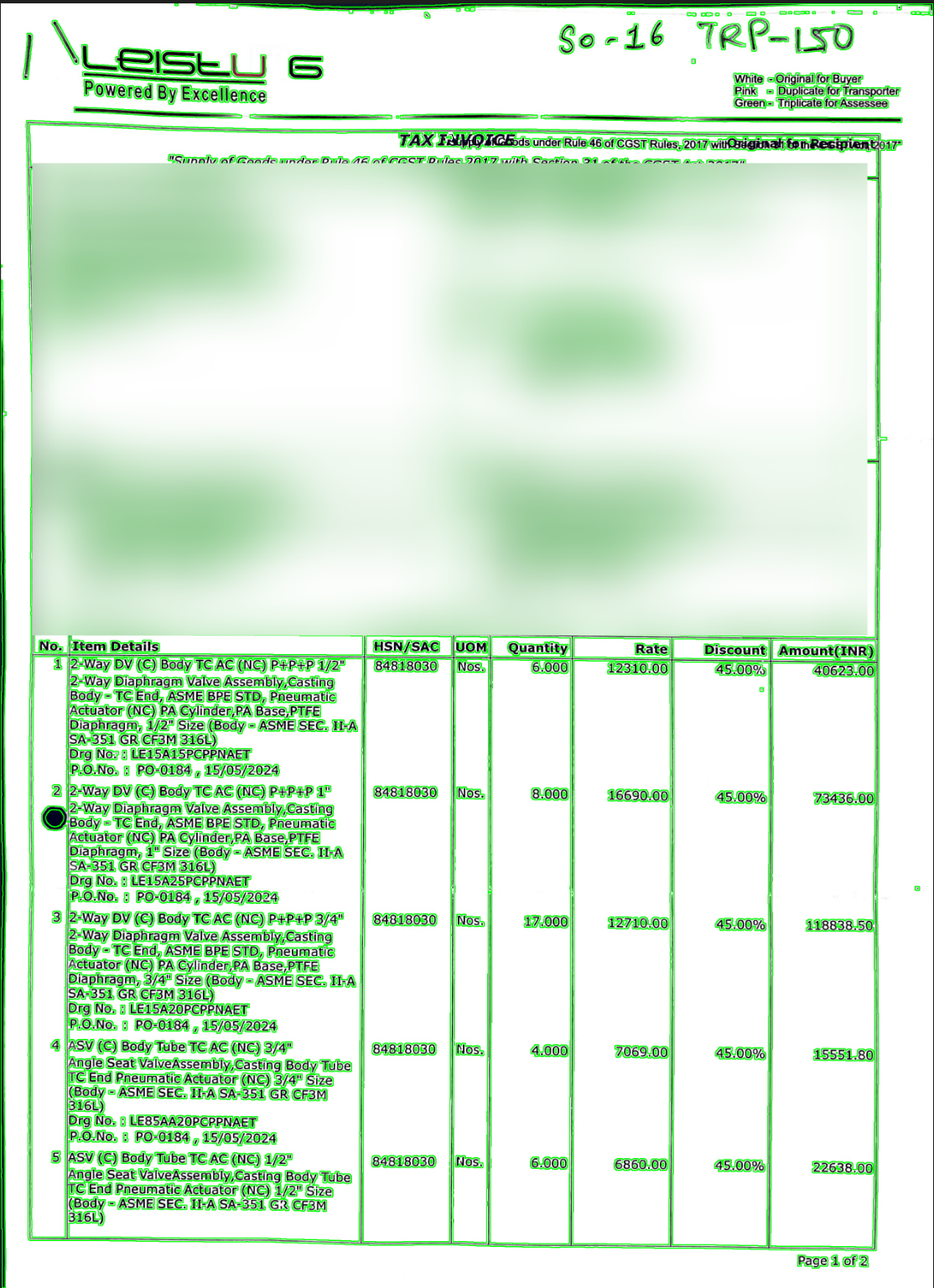}}
\hfill
\subfloat[Biggest Contour]{\includegraphics[width=0.3\textwidth]{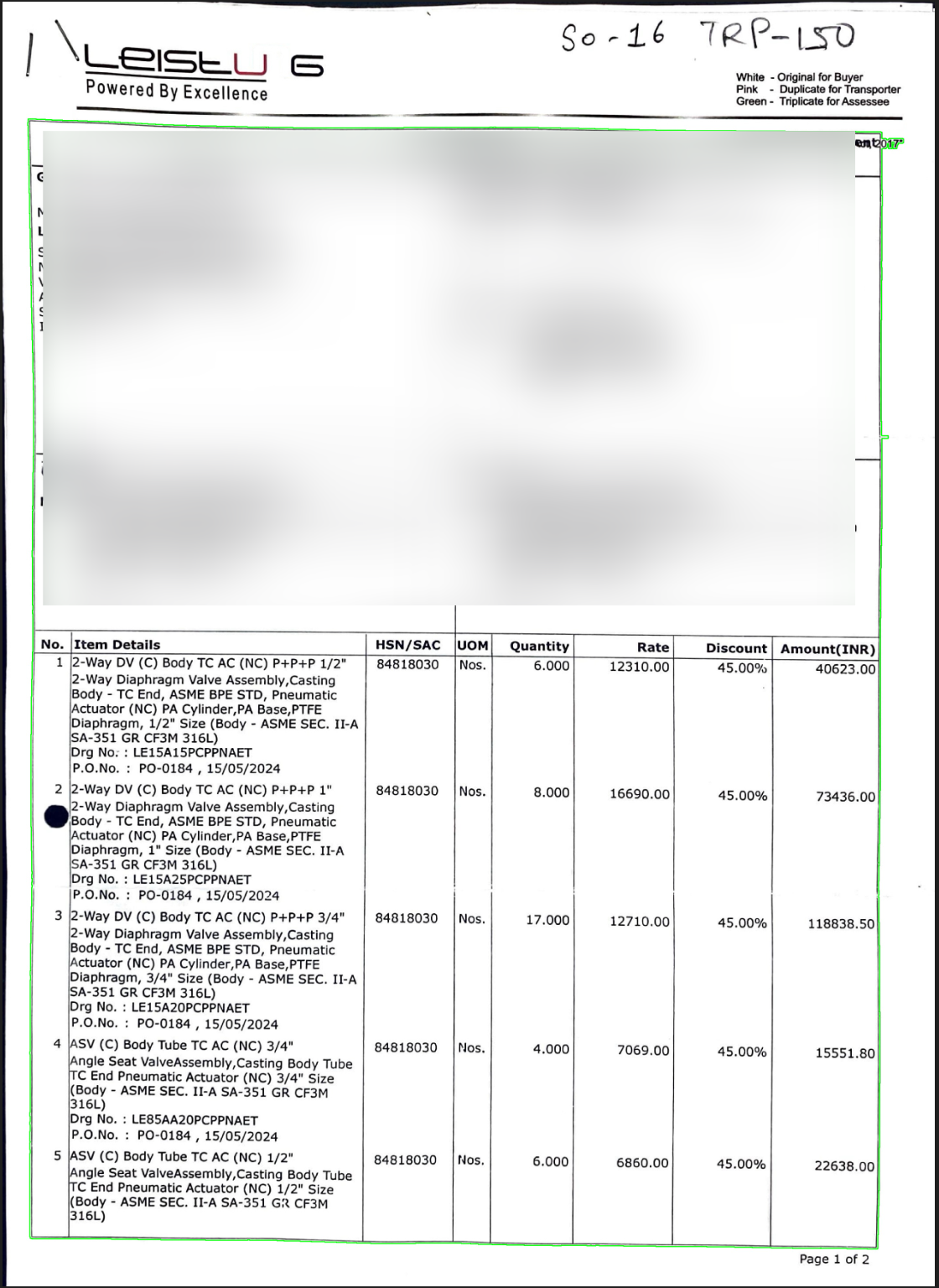}}
\par
\vspace{0.3cm}
\subfloat[Four Point Coordinates]{\includegraphics[width=0.3\textwidth]{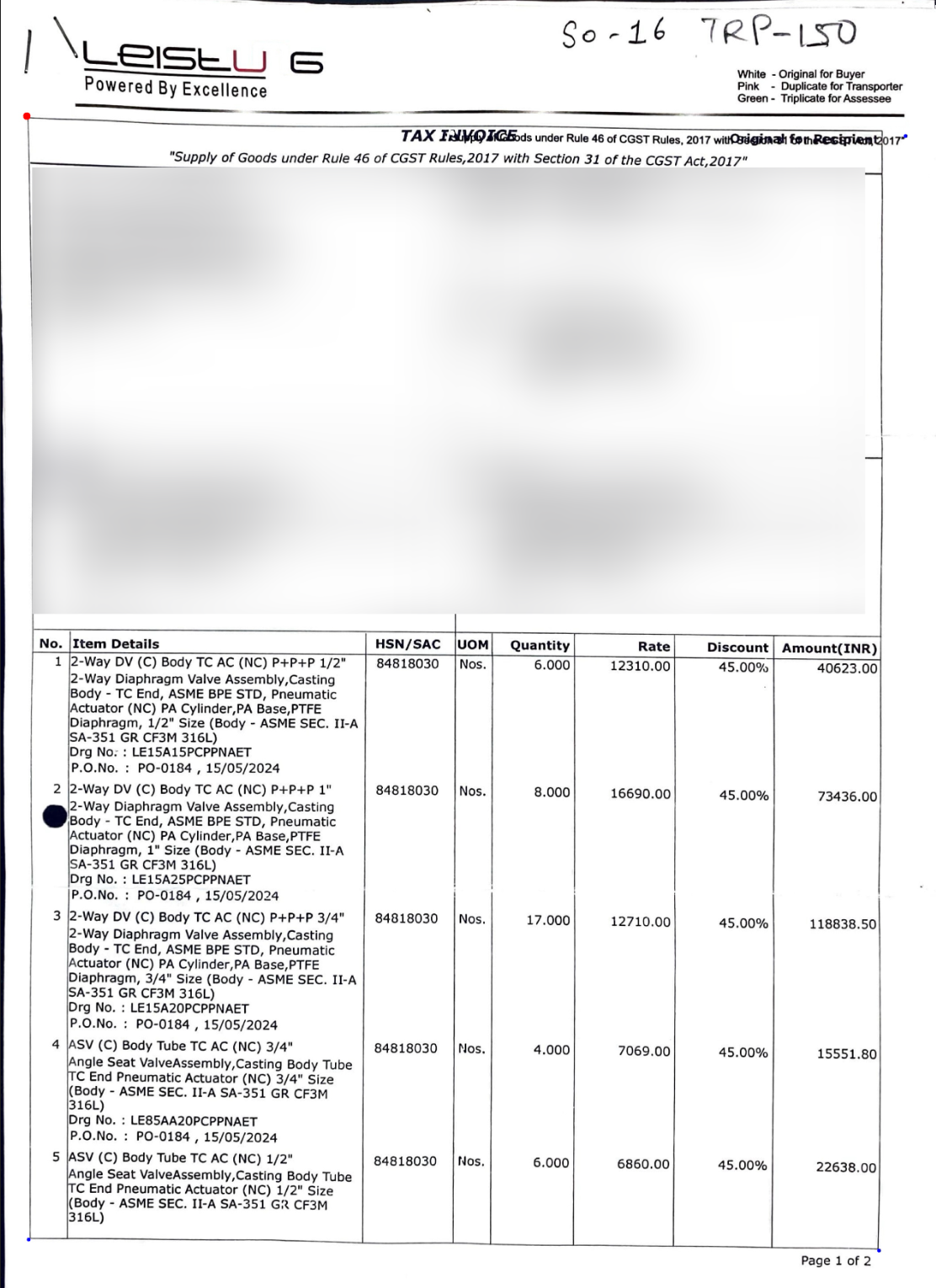}}
\hfill
\subfloat[Perspective Corrected Image]{\includegraphics[width=0.3\textwidth]{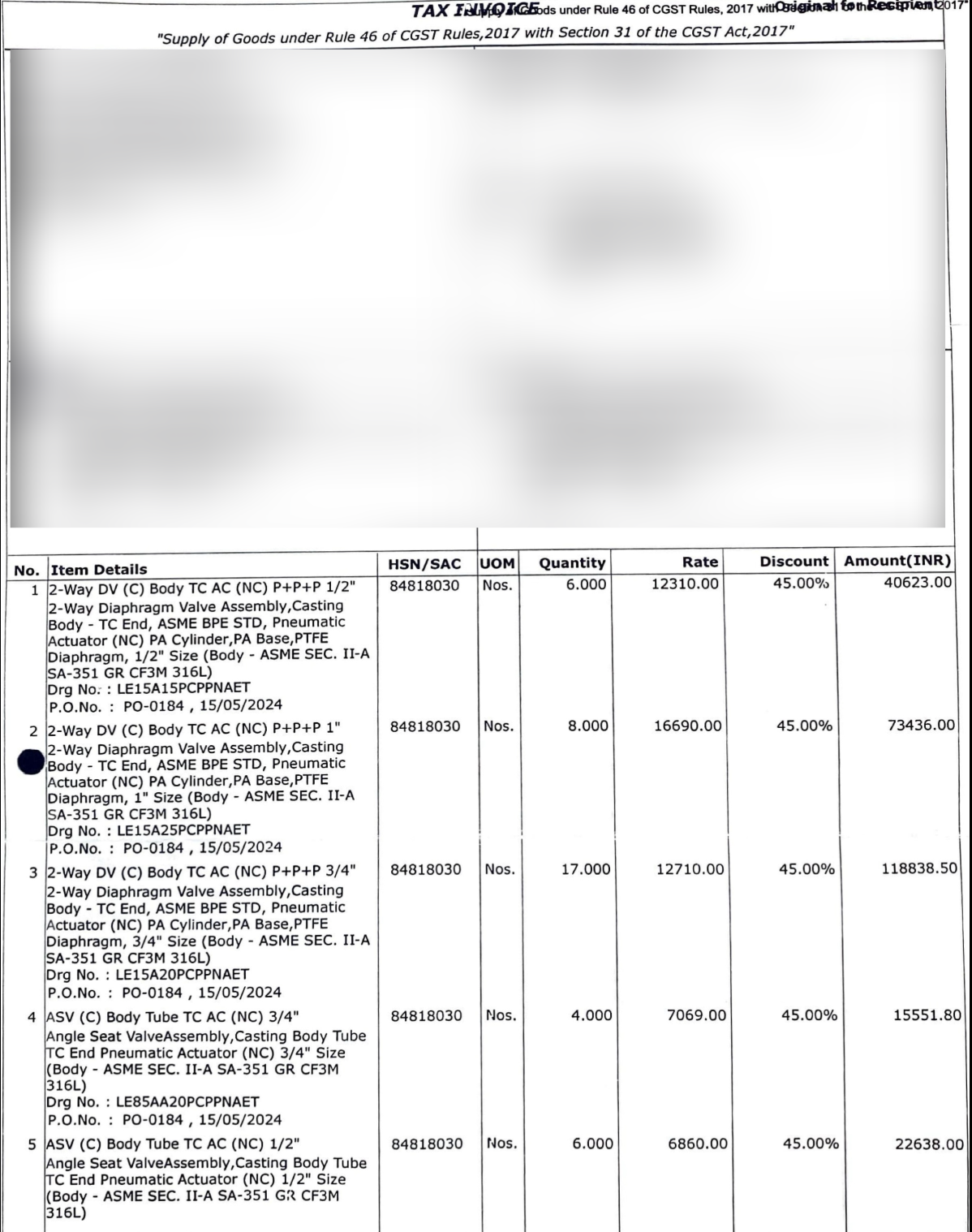}}

\caption{Image Preprocessing and Perspective Correction Pipeline}
\label{fig:preprocessing}
\end{figure}

Skewed or misaligned captures are corrected using contour-based perspective transformation. Figure~\ref{fig:preprocessing} displays this sequence.

\subsection{Table Header Detection}

\begin{figure}[H]
\centering
\subfloat[Detected Header Fields]{\includegraphics[width=0.45\textwidth]{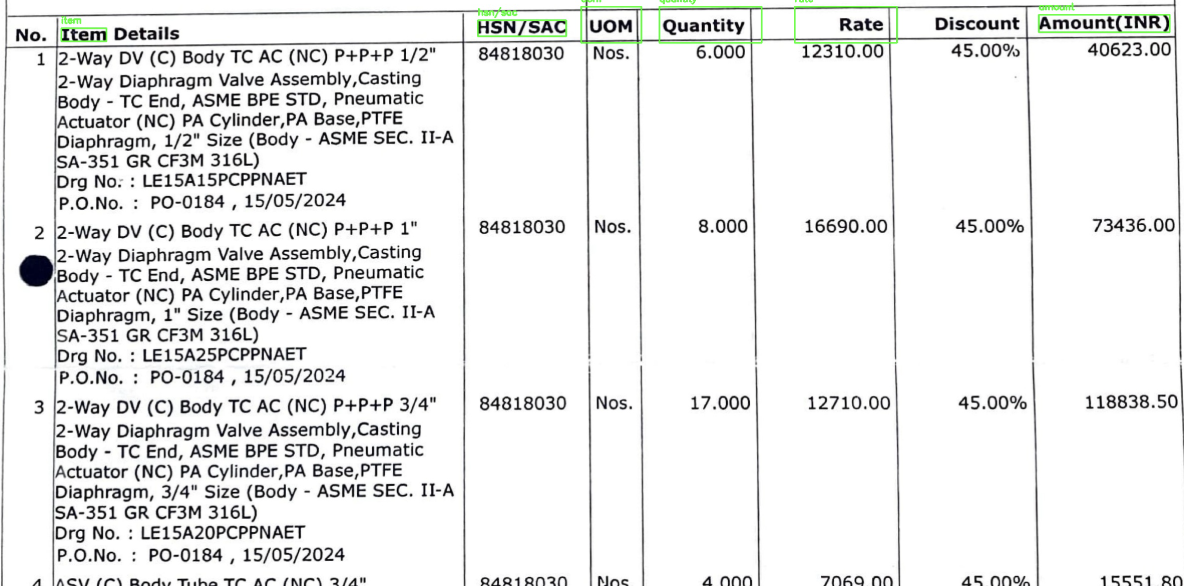}}
\hfill
\subfloat[Cropped Header Table]{\includegraphics[width=0.45\textwidth]{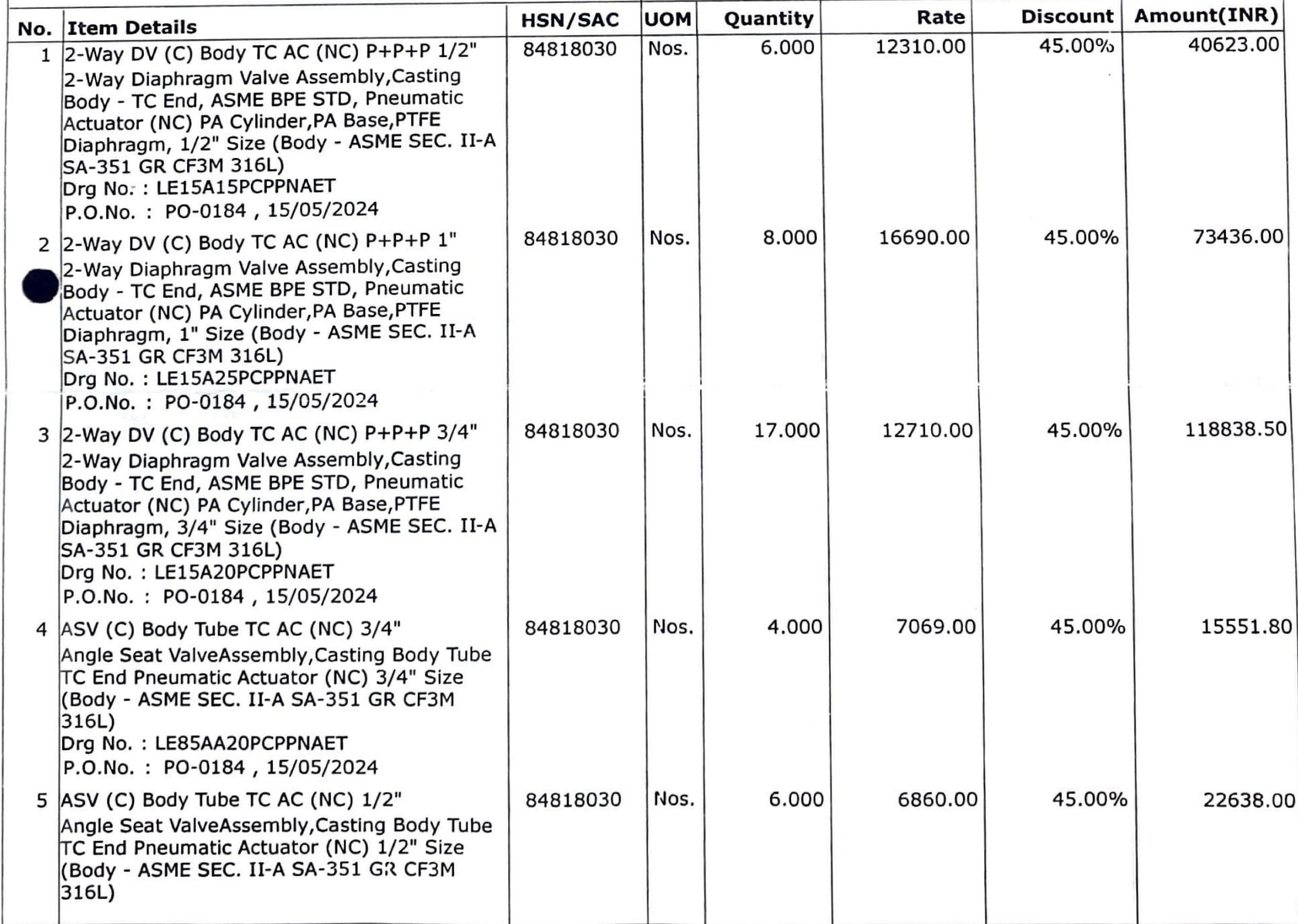}}

\caption{Header Field Detection and Cropping}
\label{fig:header}
\end{figure}

Figure~\ref{fig:header} shows extracted header sections, which typically contain metadata like invoice number, seller/buyer info, and date.

\subsection{Signature and Tick Removal}

\begin{figure}[H]
\centering
\includegraphics[width=0.5\textwidth]{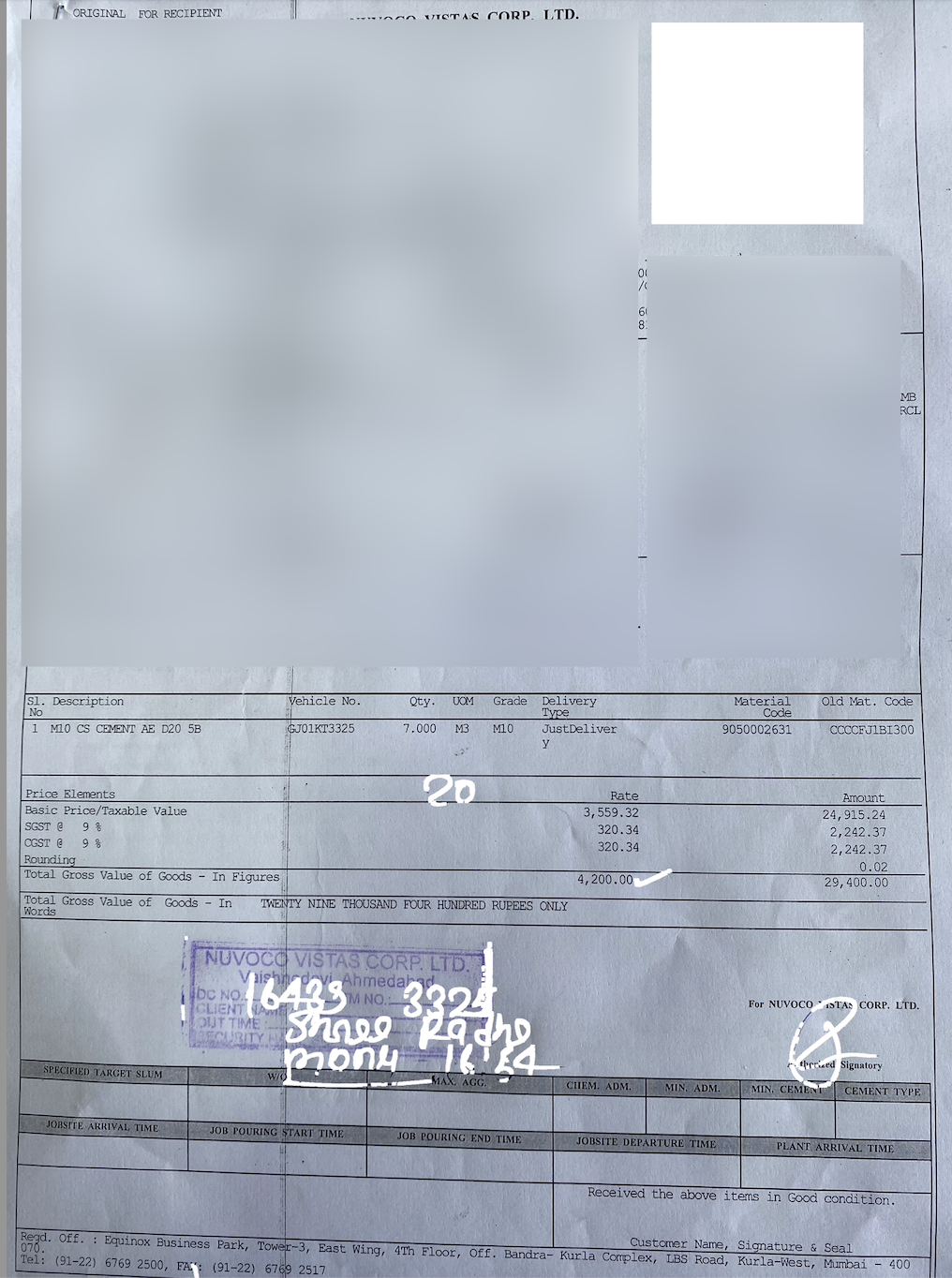}
\caption{Signature and Tick Removal}
\label{fig:signature}
\end{figure}

Ink-based user marks such as signatures or ticks may interfere with text detection. Figure~\ref{fig:signature} illustrates how such marks are identified and removed.

These elements are segmented using ink color and shape irregularities, then masked to restore the document.

\subsection{Barcode Removal}

\begin{figure}[H]
\centering
\subfloat[Before Barcode Removal]{\includegraphics[width=0.45\textwidth]{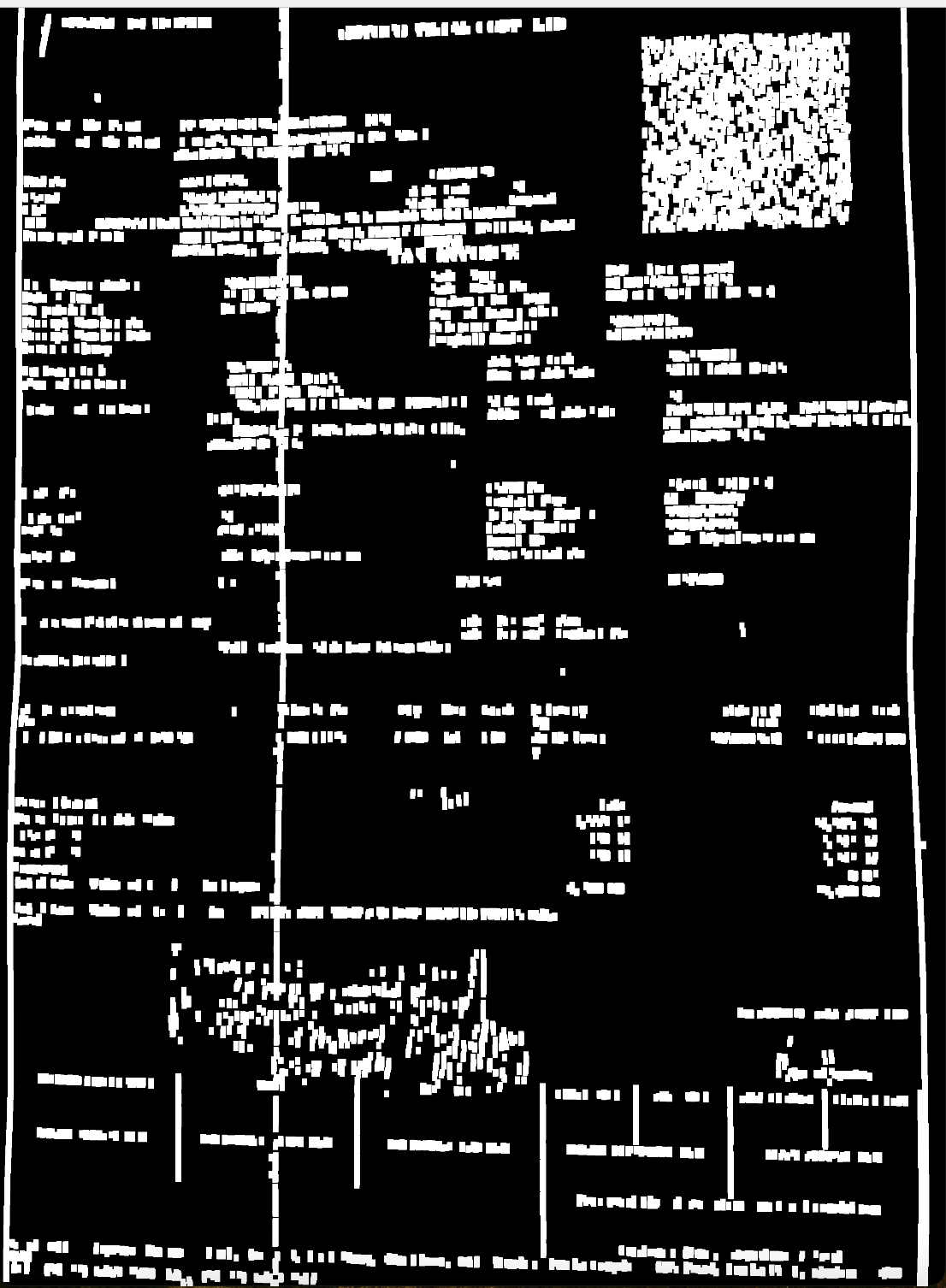}}
\hfill
\subfloat[After Barcode Removed]{\includegraphics[width=0.45\textwidth]{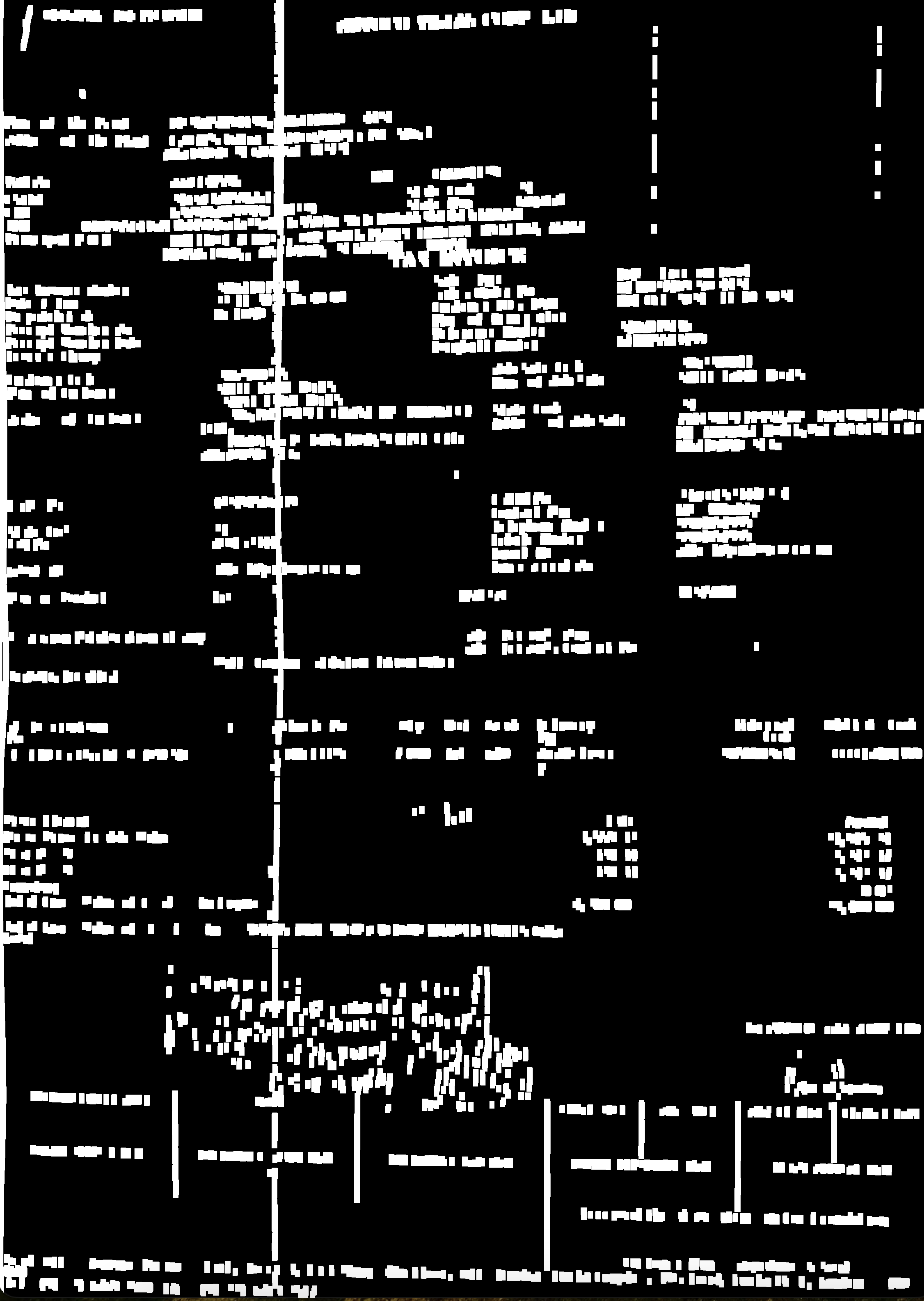}}

\caption{Barcode Detection and Removal}
\label{fig:barcode}
\end{figure}

Barcodes are identified and removed prior to OCR to avoid misclassification. See Figure~\ref{fig:barcode}.

A morphological approach isolates and eliminates these zones based on their rectangular and densely-striped nature.

\subsection{Line Removal from Product Table}

\begin{figure}[H]
\centering
\subfloat[Grayscale Image]{\includegraphics[width=0.3\textwidth]{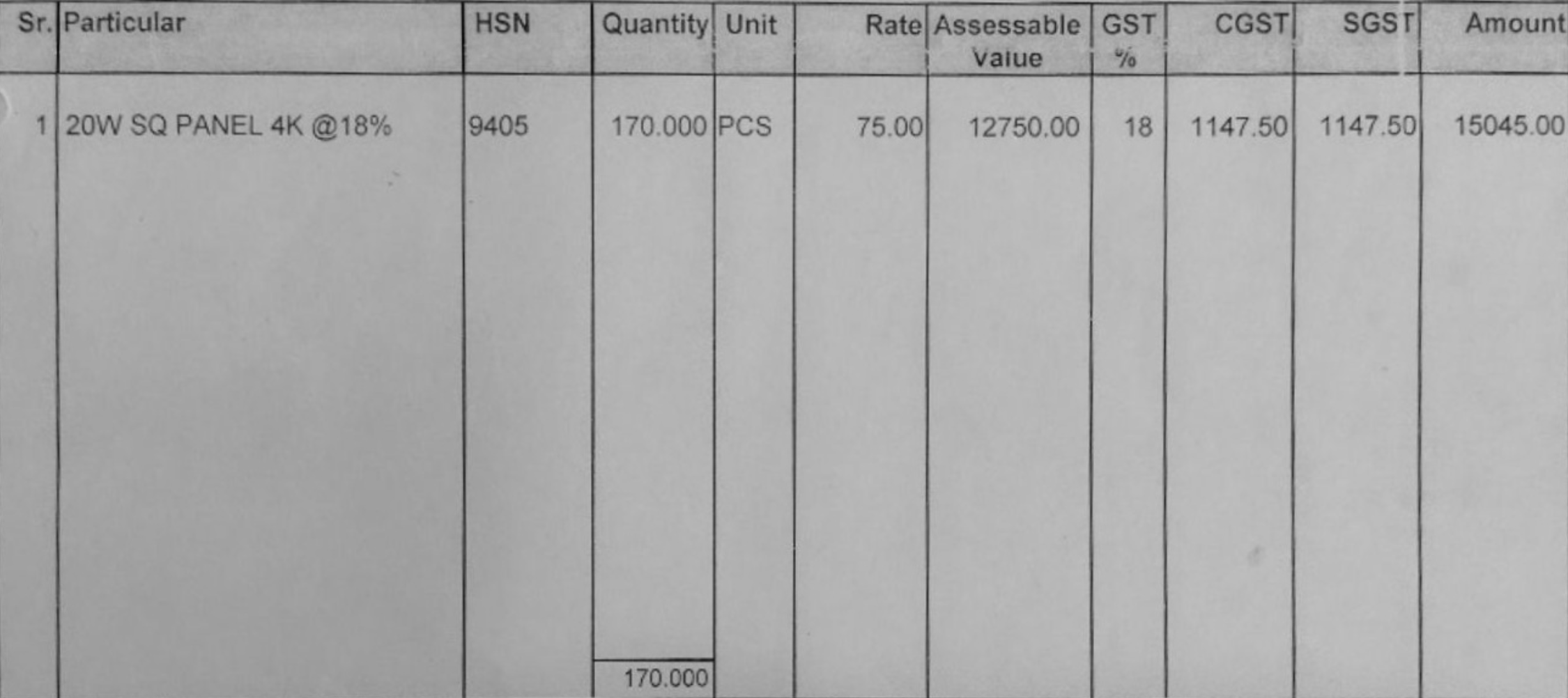}}
\hfill
\subfloat[Thresholded Image]{\includegraphics[width=0.3\textwidth]{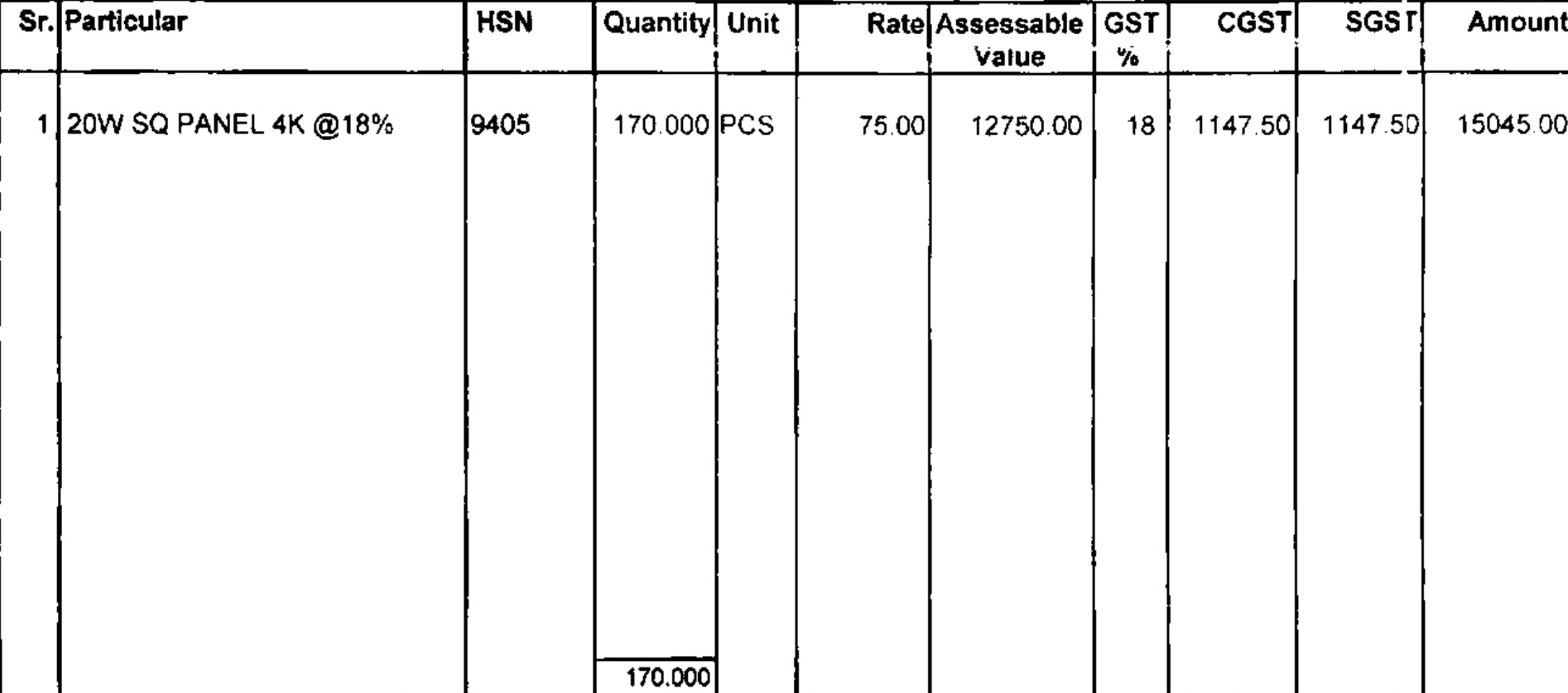}}
\hfill
\subfloat[Vertical Lines Removed]{\includegraphics[width=0.3\textwidth]{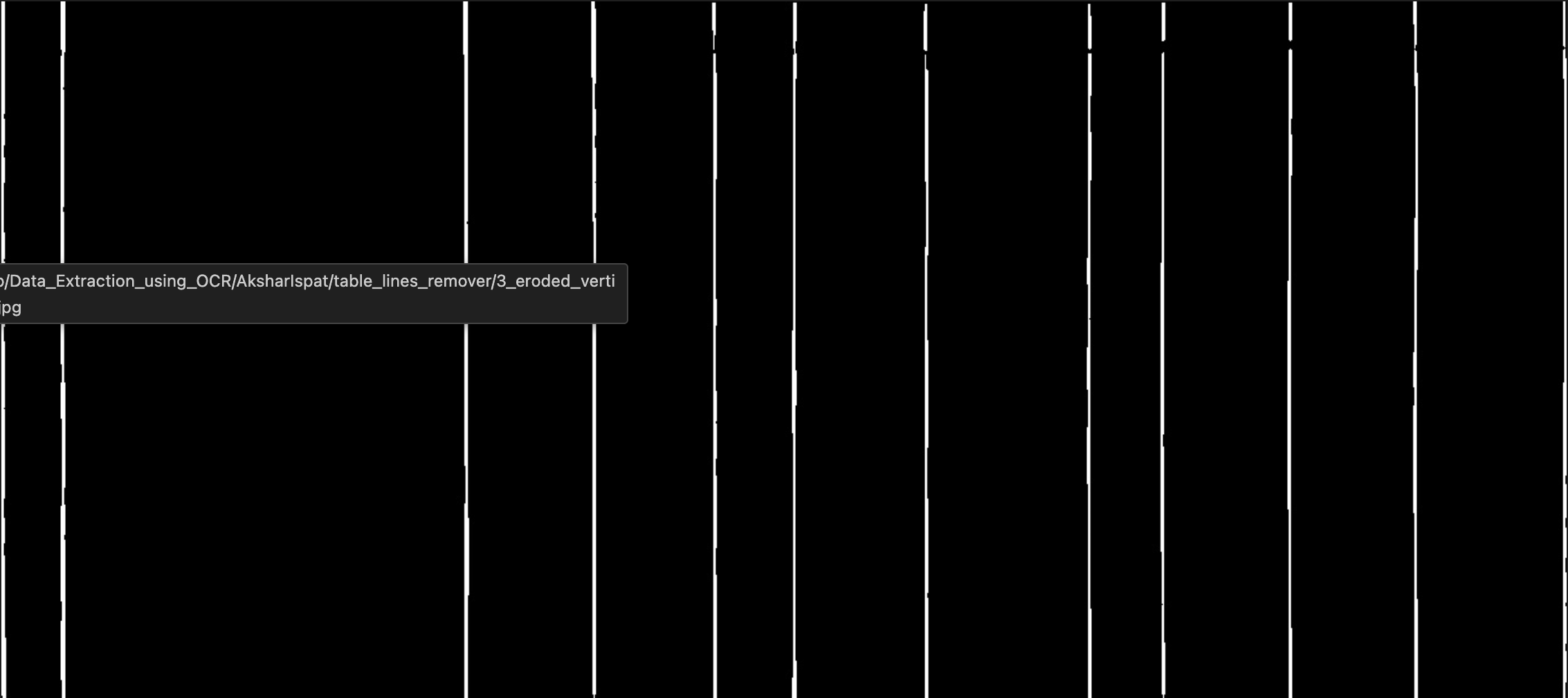}}
\par
\vspace{0.3cm}
\subfloat[Horizontal Lines Removed]{\includegraphics[width=0.3\textwidth]{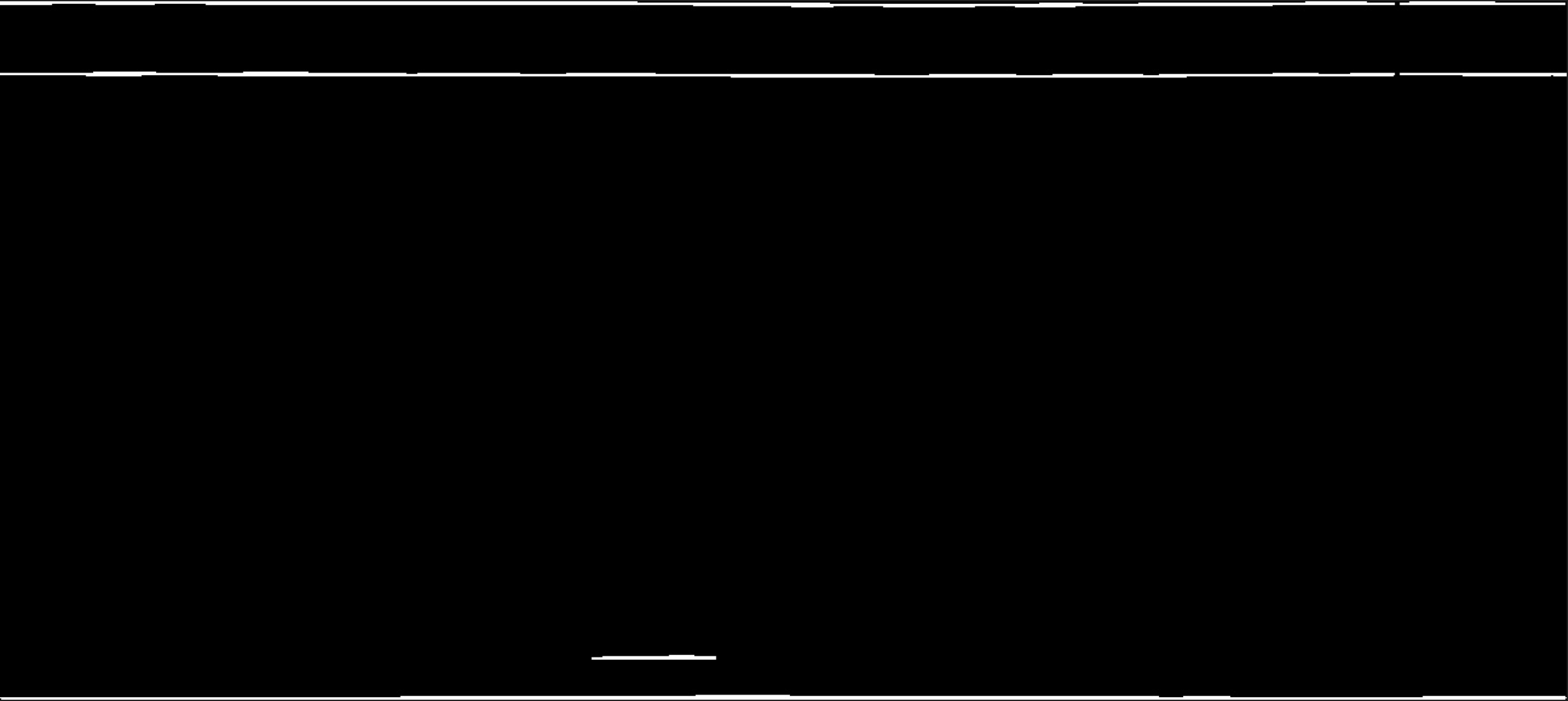}}
\hfill
\subfloat[Combined Line Mask]{\includegraphics[width=0.3\textwidth]{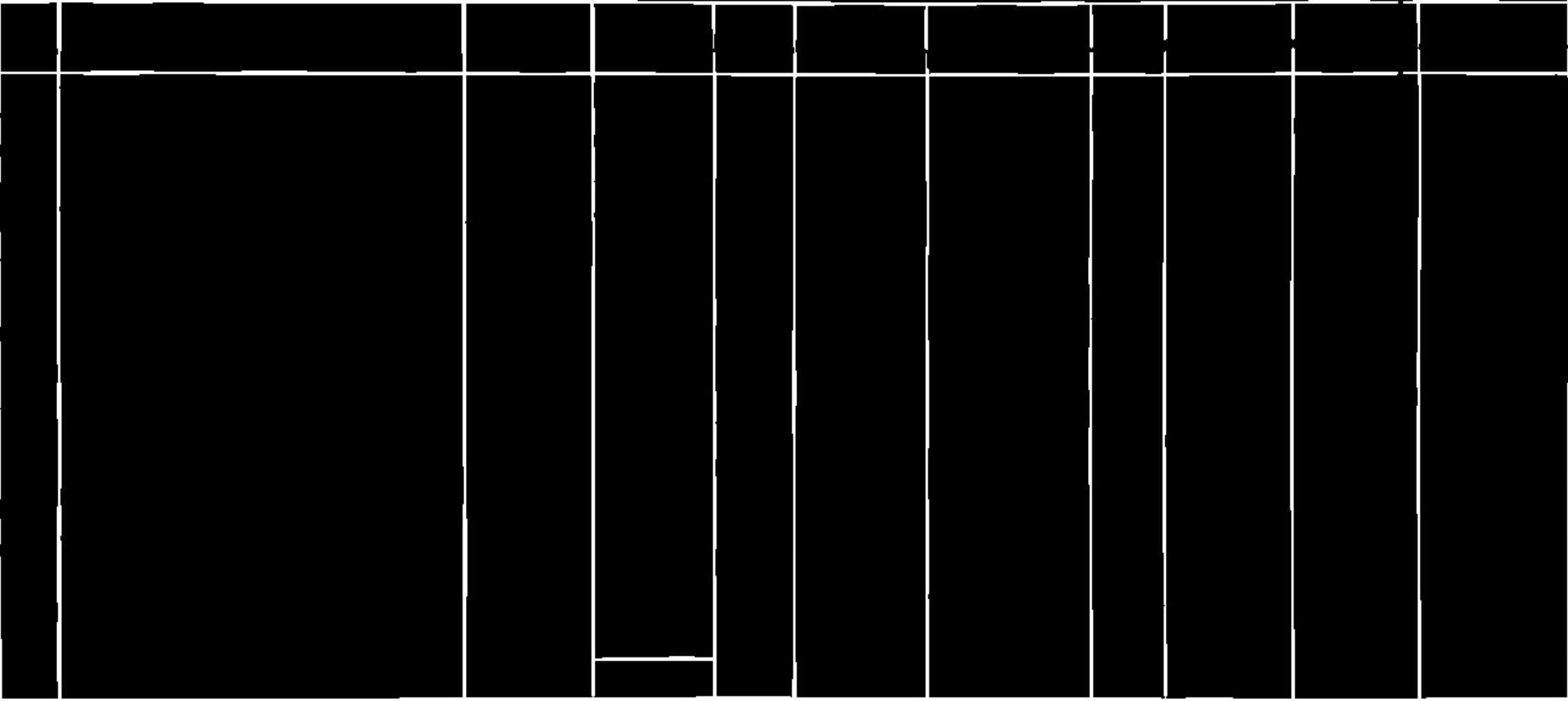}}
\hfill
\subfloat[Dilated Mask]{\includegraphics[width=0.3\textwidth]{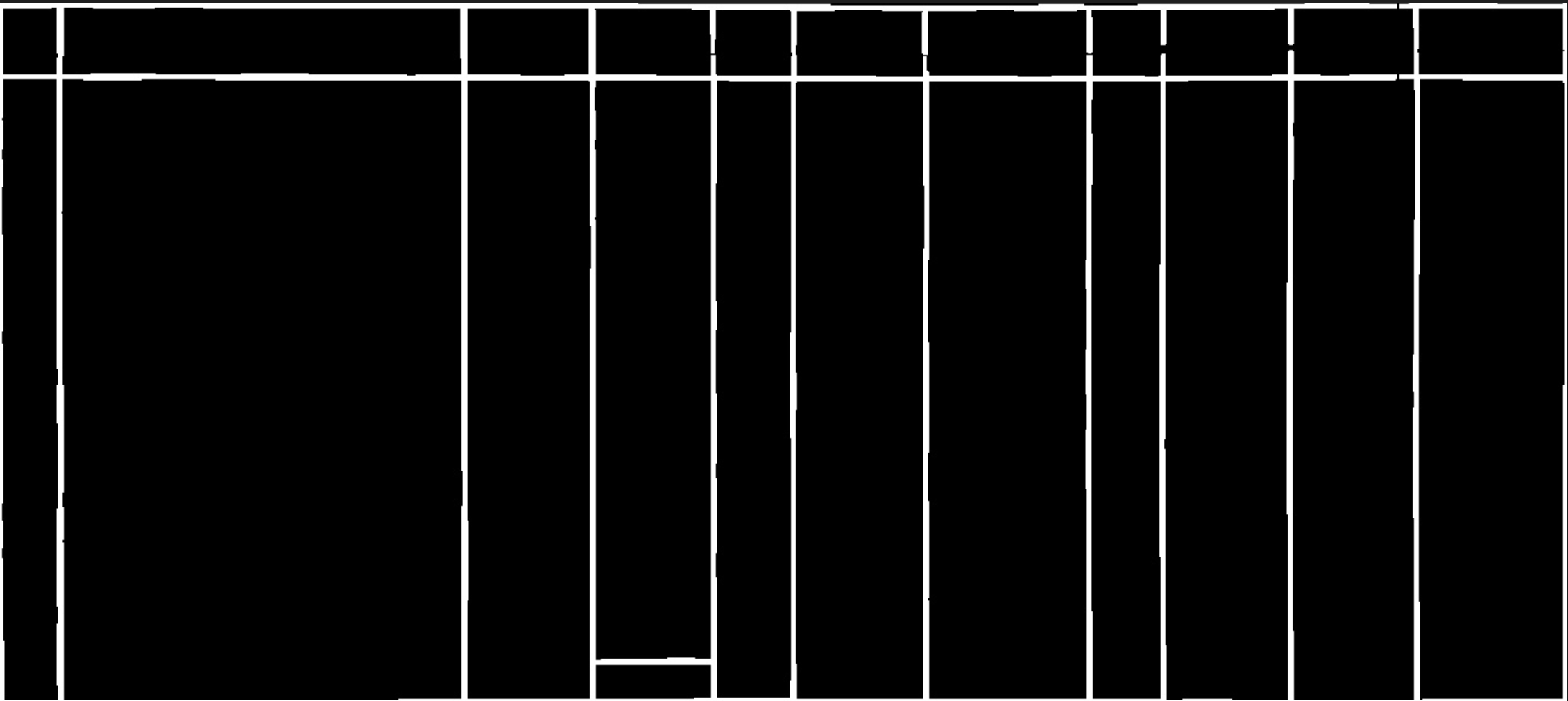}}
\par
\vspace{0.3cm}
\subfloat[Final Output - Lines Removed]{\includegraphics[width=0.3\textwidth]{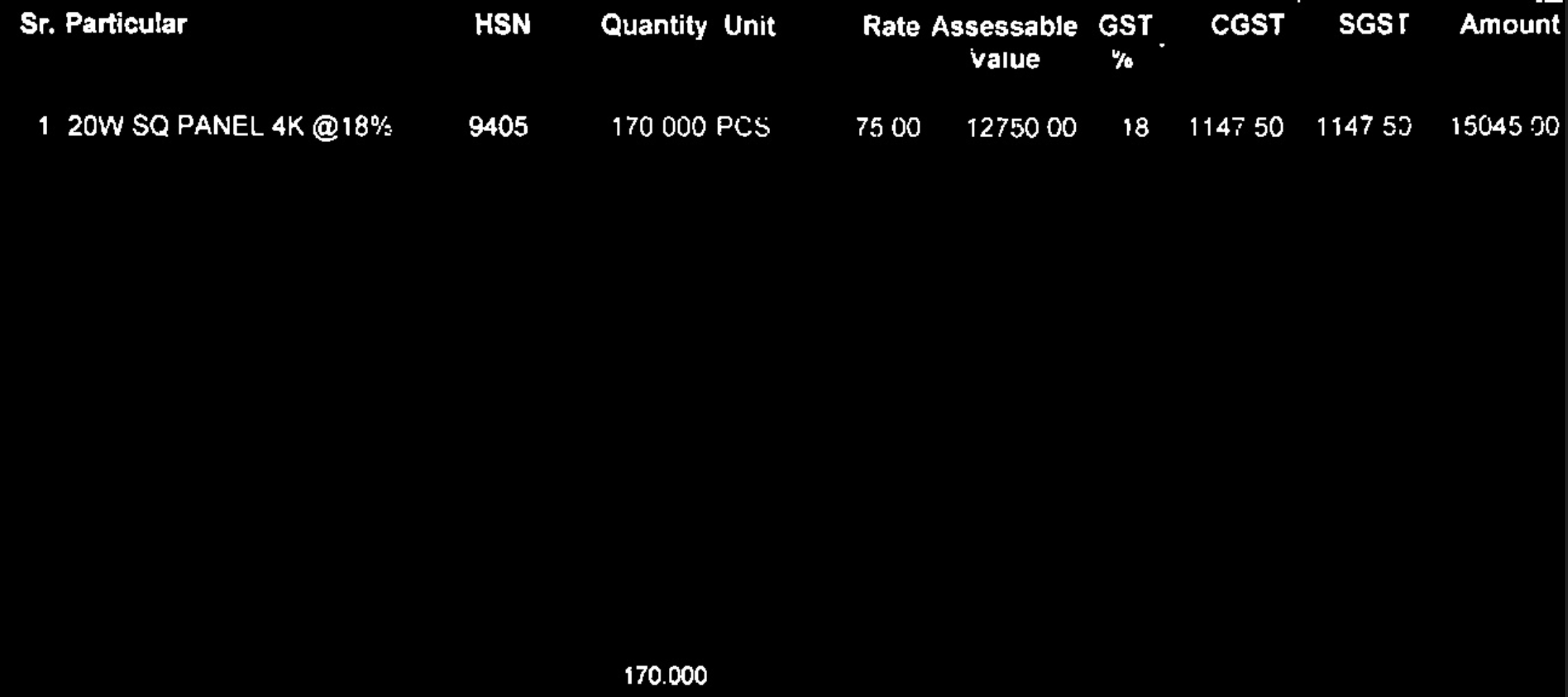}}

\caption{Table Line Removal Pipeline}
\label{fig:lineremoval}
\end{figure}

Grid lines are removed using morphological operations to isolate pure textual content. Steps are shown in Figure~\ref{fig:lineremoval}.

\subsection{OCR and Bounding Box Detection on Table}

\begin{figure}[H]
\centering
\subfloat[Dilated Table Image]{\includegraphics[width=0.3\textwidth]{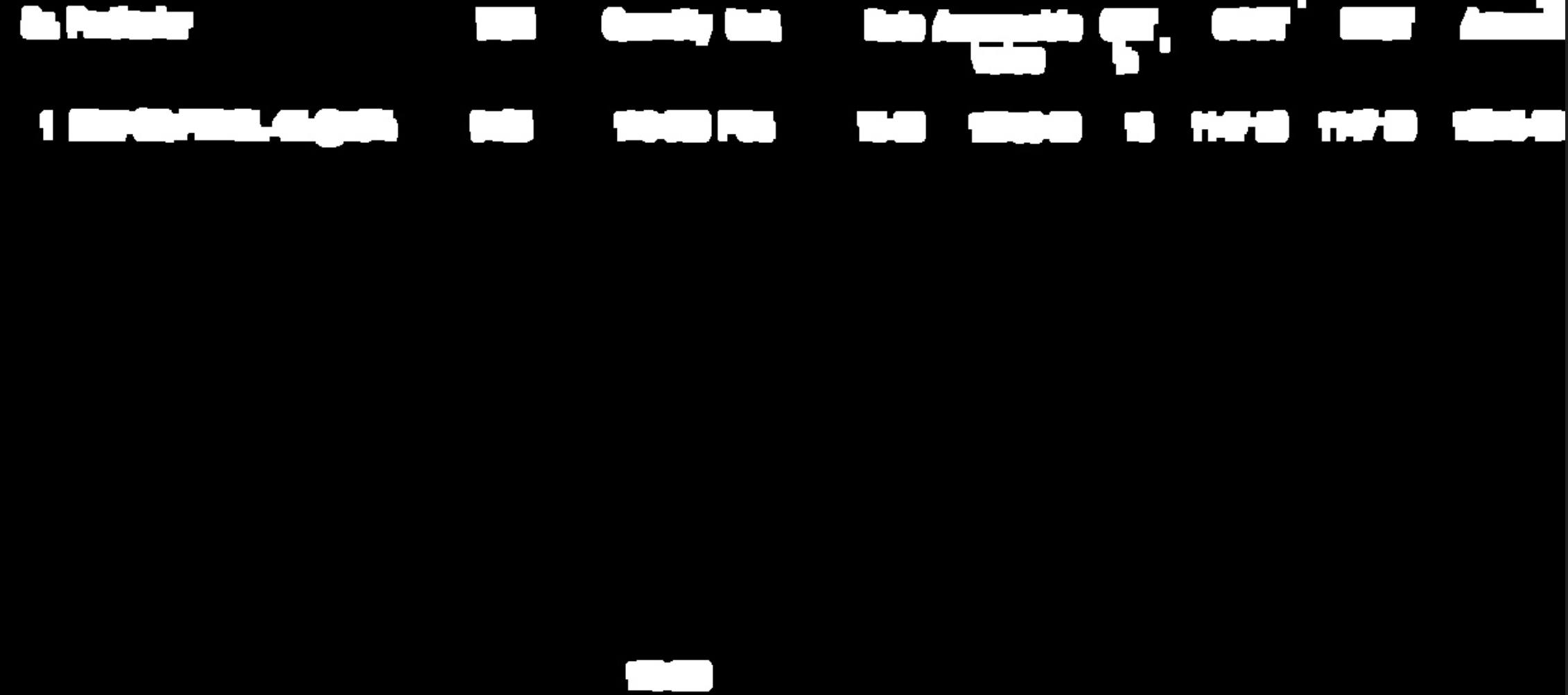}}
\hfill
\subfloat[Detected Contours]{\includegraphics[width=0.3\textwidth]{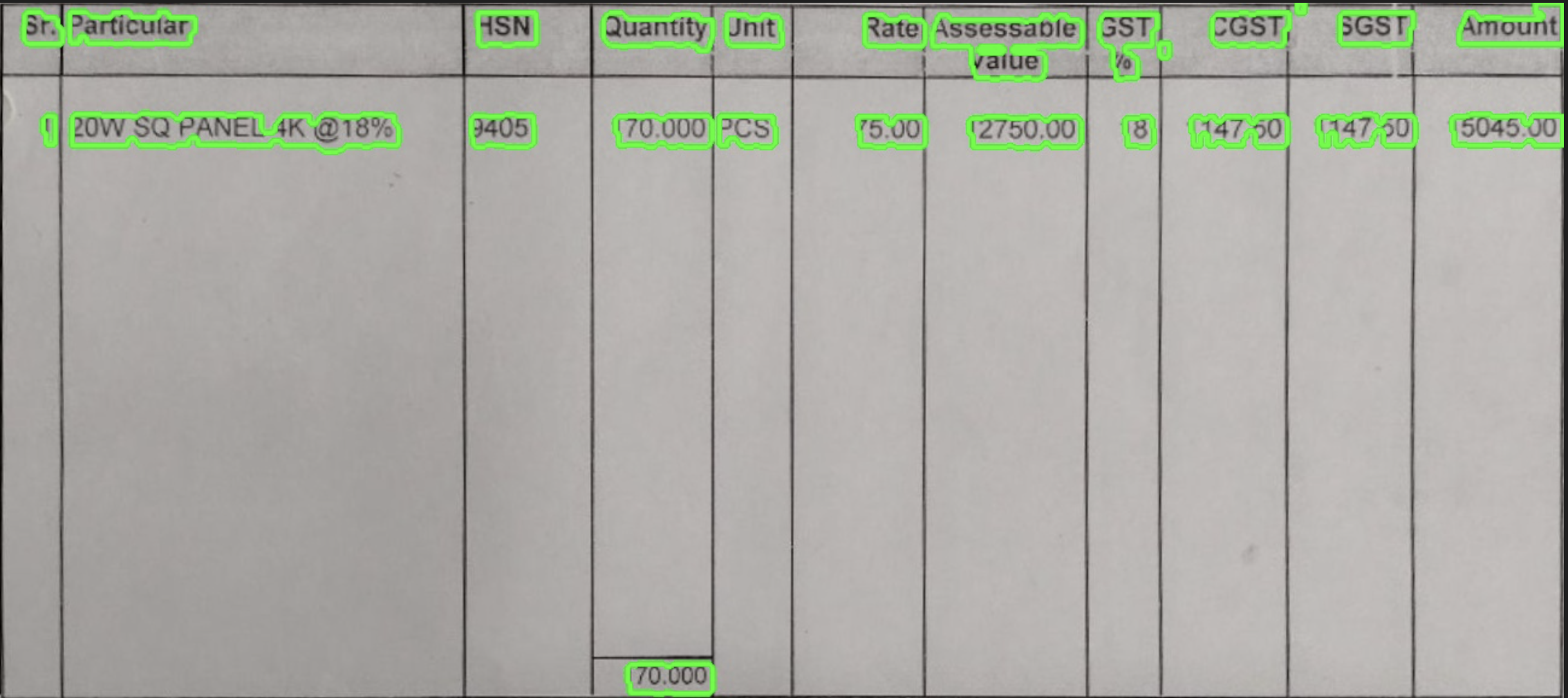}}
\hfill
\subfloat[Bounding Boxes]{\includegraphics[width=0.3\textwidth]{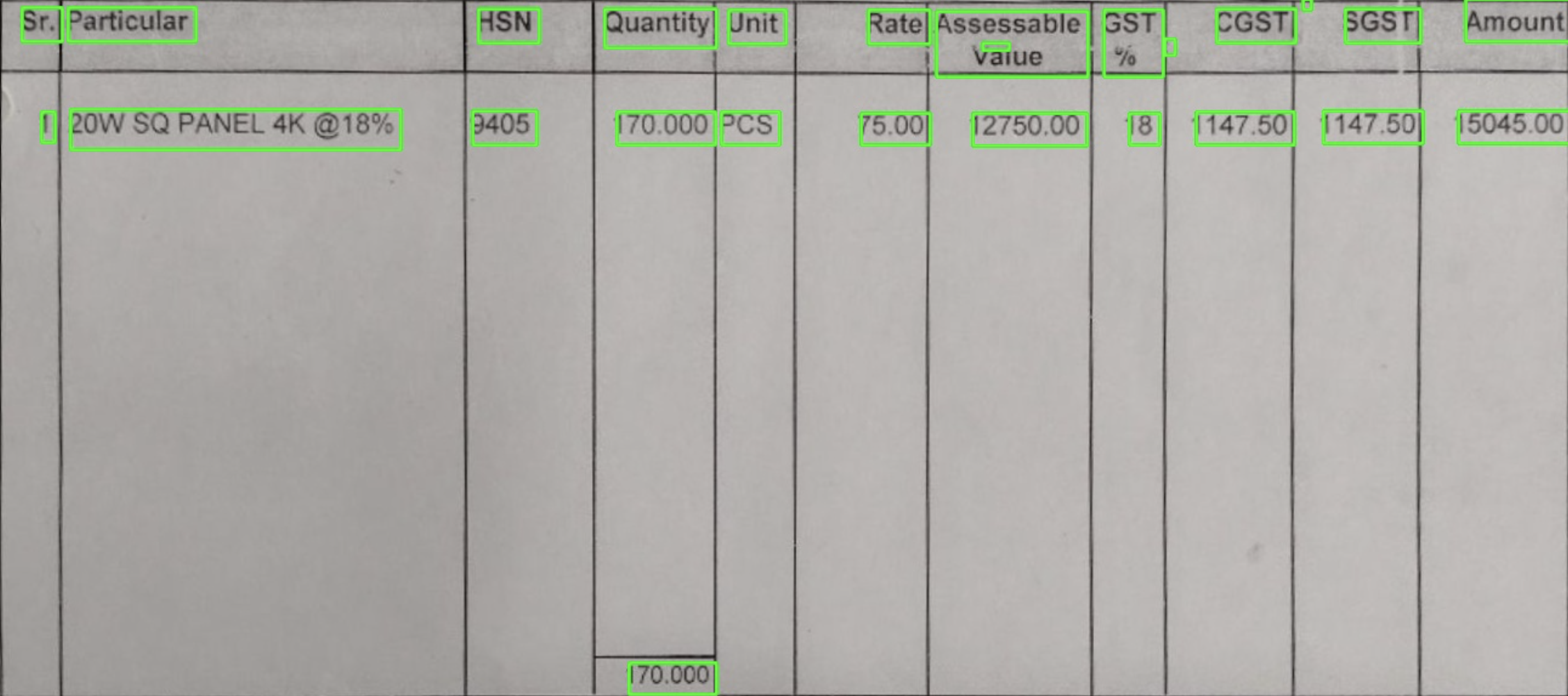}}
\par
\vspace{0.3cm}
\subfloat[OCR Result - Header Slice]{\includegraphics[width=0.3\textwidth]{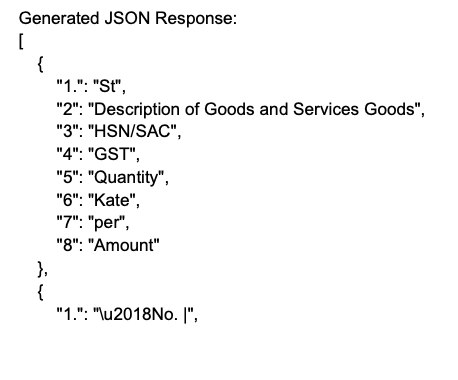}}
\hfill
\subfloat[OCR Result - Value Slice]{\includegraphics[width=0.3\textwidth]{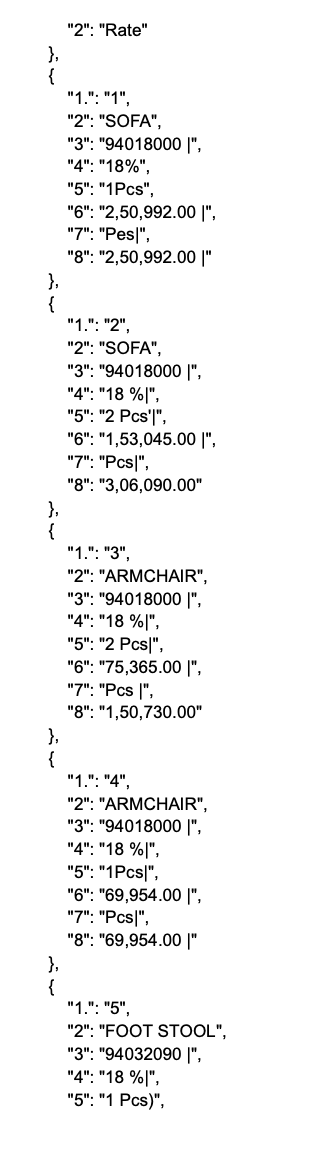}}

\caption{OCR and Bounding Box Extraction}
\label{fig:ocrpipeline}
\end{figure}

Contours and bounding boxes are extracted to segment each row and pass them to the OCR engine. Figure~\ref{fig:ocrpipeline} illustrates this.

\subsection{Final Observations}

While actual OCR accuracy figures are not shared due to private content, qualitative results show high consistency in text block extraction. Blurring the final outputs avoids exposing sensitive information, but the segmentation and processing logic demonstrates reliable operation across diverse invoice formats.

\paragraph{Conclusion of Results.}
The proposed pipeline successfully extracts key-value data from invoice images, particularly those that follow a structured tabular format with visible borders and standard alignment. The use of image preprocessing, contour detection, perspective transformation, and clustering enables reliable detection of tabular regions and separation of multi-column headers.

The system performs well on invoices with:
\begin{itemize}
    \item Clearly defined borders and grid lines.
    \item Vertically aligned columns such as \texttt{HSN}, \texttt{Qty}, \texttt{Amount}, and \texttt{Product}.
    \item Header fields placed in a consistent row.
\end{itemize}

While the pipeline is optimized for structured formats, further enhancements are planned to extend support for loosely formatted invoices, skewed headers, and low-quality scans. This modular approach allows adaptation to various invoice layouts commonly encountered in business environments.

\section{Conclusion and Future Work}

This work presents a robust and modular pipeline for extracting structured data from invoice images using traditional image processing and OCR-based techniques. By systematically addressing real-world challenges such as skew correction, table line removal, barcode and signature cleaning, and OCR noise, the proposed solution demonstrates consistent results across diverse invoice formats. The integration of tools like \texttt{img2table} and various preprocessing techniques enables effective parsing of headers and tabular content with minimal supervision.

The results show that the pipeline is particularly effective for invoices with well-defined borders, consistent header rows, and vertically aligned tabular fields such as \texttt{HSN}, \texttt{Qty}, \texttt{Amount}, and \texttt{Product}. The use of morphological operations, perspective correction, and strategic OCR slicing significantly improves downstream text recognition quality.

\subsection*{Future Work}

While the current system performs well on standardized invoice formats, several areas for improvement remain:

\begin{itemize}
    \item \textbf{Adaptive Thresholding and Erosion for Fine Text:} Current preprocessing methods occasionally merge small fonts and closely spaced characters. Future iterations will incorporate adaptive thresholding and erosion strategies tailored for finer text to preserve character separation.
    
    \item \textbf{Support for Loosely Structured Documents:} Expanding the system to handle invoices with irregular layouts, misaligned fields, and absence of clear grid lines will make it more universally applicable.
    
    \item \textbf{Dynamic Parameter Tuning:} Automating threshold, kernel, and dilation values based on the image resolution and contrast will help generalize performance across different scan qualities.
    
    \item \textbf{Confidence-Based OCR Post-Processing:} Future versions could incorporate confidence scores from OCR outputs and use fuzzy matching or language models to improve key-value mapping accuracy.
    
    \item \textbf{Integration with Machine Learning Models:} Incorporating deep learning-based table detection or form understanding models (e.g., LayoutLM or Donut) may enhance performance on noisy or handwritten invoices.
\end{itemize}

In conclusion, the presented work lays a strong foundation for semi-automated document parsing in business environments, and ongoing enhancements aim to make it more adaptable, scalable, and intelligent for practical deployment.

\end{document}